\documentclass[twocolumn, switch]{article} % Method A for two-column formatting

\usepackage{preprint}

\usepackage[numbers,square]{natbib}
\usepackage[utf8]{inputenc}	% allow utf-8 input
\usepackage[T1]{fontenc}	% use 8-bit T1 fonts
\usepackage{xcolor}		% colors for hyperlinks
\usepackage[colorlinks = true,
            linkcolor = purple,
            urlcolor  = blue,
            citecolor = cyan,
            anchorcolor = black]{hyperref}	% Color links to references, figures, etc.
\usepackage{booktabs} 		% professional-quality tables
\usepackage{nicefrac}		% compact symbols for 1/2, etc.
\usepackage{microtype}		% microtypography
\usepackage{lineno}		% Line numbers
\usepackage{float}			% Allows for figures within multicol
\usepackage{newfloat}
\DeclareFloatingEnvironment[name={Supplementary Figure}]{suppfigure}
\usepackage{sidecap}
\sidecaptionvpos{figure}{c}

\usepackage{titlesec}
\titlespacing\section{0pt}{12pt plus 3pt minus 3pt}{1pt plus 1pt minus 1pt}
\titlespacing\subsection{0pt}{10pt plus 3pt minus 3pt}{1pt plus 1pt minus 1pt}
\titlespacing\subsubsection{0pt}{8pt plus 3pt minus 3pt}{1pt plus 1pt minus 1pt}

\usepackage{booktabs}
\usepackage{mathtools, nccmath}
\usepackage{makecell}
\usepackage{subcaption}

\title{Evolutionary Tabletop Game Design: A Case Study in the Risk Game}

\usepackage{authblk}

\author[1]{Lana Bertoldo Rossato}
\author[1]{Leonardo Boaventura Bombardelli}
\author[1]{Anderson Rocha Tavares}

\affil[1]{Universidade Federal do Rio Grande do Sul}
\begin{document}

\twocolumn[ % Method A for two-column formatting
  \begin{@twocolumnfalse} % Method A for two-column formatting
  
\maketitle

\begin{abstract}
  Creating and evaluating games manually is an arduous and laborious task. Procedural content generation can aid by creating game artifacts, but usually not an entire game. Evolutionary game design, which combines evolutionary algorithms with automated playtesting, has been used to create novel board games with simple equipment; however, the original approach does not include complex tabletop games with dice, cards, and maps. 
  This work proposes an extension of the approach for tabletop games, evaluating the process by generating variants of Risk, a military strategy game where players must conquer map territories to win. We achieved this using a genetic algorithm to evolve the chosen parameters, as well as a rules-based agent to test the games and a variety of quality criteria to evaluate the new variations generated.
  Our results show the creation of new variations of the original game with smaller maps, resulting in shorter matches. Also, the variants produce more balanced matches, maintaining the usual drama. We also identified limitations in the process, where, in many cases, where the objective function was correctly pursued, but the generated games were nearly trivial. This work paves the way towards promising research regarding the use of evolutionary game design beyond classic board games. 
  %The proposed evolutionary design pipeline is capable of creating and evolving a tabletop game with different equipment, in addition to generating better versions of the Risk game.
\end{abstract}

 % (optional)
\vspace{0.35cm}

  \end{@twocolumnfalse} % Method A for two-column formatting
] % Method A for two-column formatting

%\begin{multicols}{2} % Method B for two-column formatting (doesn't play well with line numbers), comment out if using method A

%%%%%%%%%%%%%%%  Main text   %%%%%%%%%%%%%%%
% \linenumbers

\section{Introduction}
%Games are highly present in our lives, from childhood to adulthood. However, games go beyond mere entertainment. Their characteristics such as computational complexity, variability of strategies, and emergent behaviors attract academic research interest \cite{Risi2020}.

Creating and testing games is a challenging and costly task. A significant number of matches are needed to determine if a game is good enough to be published, in terms of balance and enjoyment. In this sense, Artificial Intelligence (AI) can aid in automating such tests, speeding up the search for balanced and enjoyable games.

Evolutionary game design \cite{browne2011evolutionary} combines evolutionary algorithms to evolve game rules, defined by a grammar, with automated playtesting with general game playing \cite{Genesereth2005ggp} algorithms. It has proven effective on board games, even creating the relatively successful commercial game Yavalath\footnote{Yavalath (\url{https://boardgamegeek.com/boardgame/33767/yavalath}) is a board game created via the Ludi framework  \cite{browne2011evolutionary}. Its rules are regarded as interesting by humans.}. However, the original evolutionary game design process is limited to traditional board games with simple equipment and does not support the inclusion of cards, dice, and maps found in tabletop games.

This work proposes extending evolutionary game design techniques to accommodate for the more complex rules and equipment of tabletop games, such as dice and cards. 
Our chances also include the replacement of genetic programming by a genetic algorithm, allowing a more aggressive exploration of the search space. 
We apply our extended evolutionary game design approach to Risk, a game where players battle to conquer all territories displayed on the map. 
In Risk, dice rolls determine the outcome of battles, adding uncertainty to the gameplay. Additionally, cards can be traded for troops, increasing the strategic possibilities of the game.
We aim to create novel and interesting variants of the game. Risk already has many different published variations of its original game, which turns it into a good tabletop game to test the automatic generation of variants through evolutionary game design.

%Although Risk has been explored in terms of game-playing agents and as an educational tool, little has been done regarding the game's design and AI. %The objective of this work is to use evolutionary game design in the creation and evaluation of new versions of the Risk game. 
% A previous study by \cite{Bombardelli2022} investigated evolutionary game design in Risk using genetic programming as the evolutionary algorithm. This work expands that investigation by adding missing elements from the original game and replacing genetic programming with a genetic algorithm to more effectively explore the search space.

To allow the evaluation of our extended evolutionary game design process, we use a simplified, two-player version of Risk. We also replaced the playtesting agent based on Monte Carlo Tree Search (MCTS) \cite{browne2012survey} with a faster and simpler rule-based agent to make playtesting feasible.
Our empirical results indicate that our approach is able to generate Risk versions with smaller maps, resulting in shorter matches. 
Also, the variants were  more balanced, maintaining the usual drama, where a losing player can hope to turn the tides of a match and win. 
However, we identified the limitation that, in many cases, the generated games had very small maps, which led to trivial games, despite the good evaluation of the objective function. 
Moreover, the rule-based agent provides limited coverage of the strategic possibilities of the generated versions.

The main contributions of this work include deepening the investigation of evolutionary game design for tabletop games. Our findings show that aiming to minimize Risk's branching factor favors the predominance of small maps. The source code of the implementation for reproducibility purposes is available\footnote{https://github.com/lanabr/Risk-Generation}.

\section{Background}
\label{sec:background}

\subsection{Evolutionary Game Design}
\label{sec:egd}

The Ludi framework \cite{browne2011evolutionary} was proposed with the aim of creating exciting and novel board games via software. To accomplish this, the system is capable of playing and evaluating newly created games. The remainder of this subsection describes the components of the Ludi framework.

The Game Description Language (GDL) defines the rules and components of the games. The GDL needs to be robust and comprehensive in order to represent known and unknown games, and it also needs to be understandable and coherent for human creators. Ludi's GDL is based on the concept of ludemes, which are units of game information that can be replicated and transformed.

The General Game Player (GGP) is capable of interpreting and playing any game described in GDL. Includes the rule parser, the game object, the play manager, which includes move scheduling and cycle detection, and the user interface. 

The Strategy Module is responsible for finding the move to be made in each round of the game. It uses advisors and policies to evaluate the current state of the game and determine the best move.

The Evaluation Module measures the evaluation criteria of the games. These criteria encompass intrinsic (based on rules and equipment) and extrinsic (based on game moves and outcomes) characteristics of the games. 

The Synthesis Module is responsible for creating new sets of rules using evolutionary algorithms. The process involves selection, crossover, mutation, and rule validations. The initial population consists of real games with rule diversity. %, and genetic variability is valued throughout the process.

The current version of the system, Ludii (two ``i'' mean ``second version'' of Ludi) \cite{Browne2018ludii}, does not yet generate new games. 
%It includes a game description language, game-playing agents, an evaluation module with fewer criteria than the original version, and a graphical user interface.
The main goal of Ludii now is to reconstruct ancient games from partial rule descriptions and equipment from archaeological findings.

\subsection{The Risk Game}
\label{sec:riskrules}
Risk is a strategy tabletop game for up to six players, simulating military conflicts in a world map\footnote{https://www.hasbro.com/common/instruct/risk.pdf}. On the basic version, a player wins by conquering all map territories (i.e. by placing at least one troop piece at each territory). 
This section briefly describes the Rules of the Risk game. At the beginning of the game, the territories are distributed evenly for each player. 
After that, each player plays a turn consisting of four steps:

\begin{itemize}
    \item Receive and place troops: players receive troops proportional to the territories and continents they own, and may be able to exchange territory cards for extra troops;
    \item Attack opponents: players may attack  neighboring territories. Both the attacker and the defender roll a die for each attacking or defending troop, at the maximum of three. Draws favor the defender. If the attacker wins by defeating all defending troops, he conquers and occupies the disputed territory;
    \item Fortify territories and receive territory card: on the aftermath of all attacks, the player may move troops to reinforce territories on enemies' borders. If the player conquered at least one territory, it receives a territory card.
    
\end{itemize}

Players alternate turns until a single player own all game territories.

\section{Related Work}
\label{sec:related}
We begin by showing that research on content generation for Risk has not been previously considered (Section \ref{sec:related-risk}). Then, Section \ref{sec:related-generation} shows that, although the automatic generation of specific content for games, such as maps, cards, sounds, and textures has been widely explored \cite{hendrikx2013, togelius2011} by researchers in both physical and digital games \cite{summerville2016mystical, uriarte2013} the generation a complete game becomes more challenging due to the need to consider important features such as rule coherence and game enjoyment.

%Some studies have used Risk games as test platforms, along with research aiming to create games through evolutionary methods. These works highlight the importance of these themes and indicate that there is still much to be researched and developed in this field.

\subsection{Risk and applications}
\label{sec:related-risk}
Much of the research on Risk focuses on game-playing agents. Different methods are employed, such as Monte Carlo Tree Search (MCTS) \cite{Brand2014}, heuristic-based approaches \cite{Olsson2005}, and Convolutional Neural Networks \cite{Carr2020}. Some works combine the use of MCTS with Neural Networks \cite{Heredia2021}, while others utilize linear evaluation functions \cite{Wolf2005}.
Although competent game-playing agents might be useful for playtesting, these works did not focus on content creation for the game. 
Moreover, the agents might not be able to cope with a game with changing rules.

%Although these methodologies have their advantages and disadvantages, they all aim to pave the way for game improvement and test various existing methods. Tree-based approaches, for example, may be limited by the evaluation function, while approaches that analyze the situation in small steps can simplify and enhance the quality of the solution.

Some studies \cite{Brand2014, Olsson2005, Carr2020} utilize Lux Delux\footnote{https://sillysoft.net/lux/}, an online platform based on Risk, to test different game profiles and AI. However, this is a commercial platform not focused on content generation, and some rules are different from the original Risk.

Other research examines more specific situations in the games, such as battle outcome estimation \cite{harju2012probabilities} or even the use of Risk or particular variants as learning tools \cite{marks1998using}. %TODO se o paper for aceito, pode adicionar a citação do War para aprendizado

% In summary, tabletop games in general and Risk in special have not been extensively explored and many research challenges remain. 
% In special, content creation and rule variation seems underexplored.
% This work presents an application of evolutionary game desing for tabletop games focusing on Risk, as an initial investigation of the potential of the approach and clarification of the associated research challenges.

%applied to  While many research efforts have focused on developing AIs to play these games, there are various techniques and combinations yet to be explored. Additionally, other metrics, such as map balancing and troop gains, can also be valued. Overall, the number of works involving Risk and War is limited when considering areas beyond those mentioned.

\subsection{Automated game generation}
\label{sec:related-generation}

In the field of automated game generation, researchers have focused on different aspects and challenges. Some works aim to generate specific game content, such as maps for video games \cite{snodgrass2017, uriarte2013} or cards for collectible card games \cite{chen2020chaos}, while preserving the overall environment and original rules. These works employ automated techniques to create game elements efficiently.

When it comes to generating an entire game, including the rules, the challenges become more complex. Early attempts, such as \citeauthor{Togelius2008}'s work, focused on evolving the rules of a game, particularly in the context of the Pac-Man game style. The objective was not to generate optimal games right away, but to demonstrate the possibility and explore new research areas.

Other approaches, like \citeauthor{Hom2007}'s, aimed to generate balanced games using evolutionary techniques. The goal was to create games without favoritism towards the player going first, and the process achieved considerable success. 

%It is important to note that these works do not aim to replace human designers, but rather to complement them. They explore new game mechanics, genres, and serve as support tools for human designers.

%The mentioned works highlight the importance of applying their methods to different games and objectives. There is a desire to automatically generate new rules, boards, and pieces, rather than relying solely on predefined ones. Human validation is ultimately the most important test, since human players are the primary stakeholders interested in new and enjoyable games.

% Building upon \cite{browne2011evolutionary}, \cite{Bombardelli2022} applied the evolutionary game design process to create new variants of the tabletop game Risk. This extension introduced additional components such as boards, dice, and cards, introducing randomness into the pipeline. The main achievement was the generation of new Risk maps, although they had limited influence on the match's quality. However, Bombardelli's work had its own limitations, such as simplified versions of Risk and limited evaluation criteria. Further improvements and evaluations were necessary to address these limitations.

The testing platform used in automated game generation plays a crucial role. Lux Delux, developed by Sillysoft, is widely utilized by several works \cite{wiklund2015evaluating, ferrari2022towards, gibson2010automated}. However, Lux Delux has limitations, particularly in rule modification, as it is not an open-source software. The TAG (Tabletop Games) framework \cite{gaina2020tag} provides a structure to encompass different types of modern games in a common platform. However, incorporating the framework into the evolutionary design process presents challenges, as metrics acquisition and game definitions differ. 

In summary, although game creation has been explored in the literature, little attention has been given to an automatic generation of tabletop games. Regarding Risk, specifically, most research focuses on game-playing agents or educational aspects of the game.

\section{Evolutionary Design of Risk Variants}
\label{sec:approach}

The Evolutionary Game Design framework has been proven successful at generating enjoyable board games for human players (see Section \ref{sec:egd}). 
%Game rules respect a grammar, and an evolutionary process, namely genetic programming \cite{Koza1994gp} generates new games by combining the rules of previous games. 
%An eva
However, the framework handles board games with simple equipment, basically the board itself, with various forms of regular connectivity (e.g. squares, hexagons), and pieces of various types. 
This allows the definition of games such as Hex, Go or Chess, but not games that use different equipment such as dice and cards or are played on environments with nontrivial connectivity, such as maps.

This section presents our evolutionary game design approach for tabletop games, containing dices, cards and a map. We focus on Risk, a military strategy game of map domination (see Section \ref{sec:riskrules}), but the principles can be applied to any tabletop game. 
The definition of a game description language for tabletop games would be very complicated to accommodate the different equipment and the nontrivial connectivity of a map, and it is out of the scope of this work.
Instead, we define basic, non-changeable game rules (Section \ref{sec:simplerisk} and expose a set of parameters that define many features of the game behavior (Section \ref{sec:parameters}).
We use a genetic algorithm as our evolutionary approach (Section \ref{sec:ga}), combining game metrics from the original evolutionary game design process into our fitness function. 
The genetic algorithm allows for more aggressive exploration of the search space as the entire population is generated every generation, whereas in the genetic programming \cite{Koza1994gp} approach of the original evolutionary game design process creates a single individual per generation.

The proposed process of automatically generating versions of Risk is thus capable of creating, playing, and evaluating the generated versions. 
The implementation was done in the Python language and is openly available on GitHub\footnote{https://github.com/lanabr/Risk-Generation}.

\subsection{Simplified Risk}
\label{sec:simplerisk}

%The allocation phase depends on a game parameter (see Section \ref{sec:parameters}). The territory card exchange was also maintained. When a player has 5 cards, they are required to perform the exchange. However, each player's cards are not kept hidden. This was done to maintain the version simple.
%During the battle phase, attacks and defenses occur as presented in the rules and also according to the chosen variation parameters. In the fortification phase, players also have complete freedom to choose the changes to be made.

In this pipeline, a simpler version of Risk was used.
Our main simplification is the restriction to two players, to allow for more direct application of the game quality metrics (see Section \ref{sec:ga}).
Moreover, the original Risk has a variant with objective cards. Our version disregards these cards, so that the goal of each player is always to conquer all map territories. %The main difference from the original version is the omission of objective cards. This type of card would disrupt the perfect information characteristic, since the objective is a private and unique element for each player. Therefore, in all possible variations, the objective of all players is to conquer all territories on the map.

\begin{figure}[!ht]
    \centering
    \includegraphics[width=0.9\columnwidth]{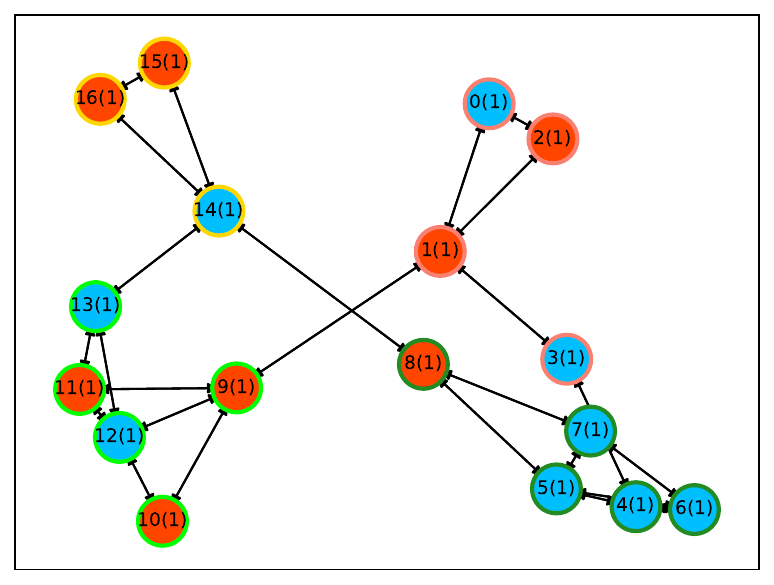}
    \caption{Map visualization on the implemented Risk engine.}
    \label{meio}
\end{figure}

The implemented game engine provides a simple visualization, as shown in Figure \ref{meio}. Circles represent territories, and lines connect neighbor territories. Nodes are numbered as ``ID(troops)''. 
Colors surrounding nodes indicate their continents, whereas color filling the nodes indicate the owner.

The game engine has a command line interface for a human player. Although primitive, the interface allows the player to perform all possible actions and play the game properly. % The instructions and commands used during a game with a human player are simple and easy to play.

\subsection{Parameters}
\label{sec:parameters}

%The differences between the games Risk and War motivated the search for details that could be changed to try to improve the game. 
The game parameters define the map and some aspects of the rules, as outlined next. We call the three first parameters as ``attributes'' and the last one is the ``map''. Attributes and map have different treatment under crossover and mutation. %In addition to the changes made to the map, such as the removal and addition of territories, already discussed by \cite{Bombardelli2022}, this work aims to customize 4 more aspects:

\begin{itemize}
    \item Initial territory distribution: flag indicating whether territories are randomly distributed or alternately selected by players;
    \item Number of defensive dice: two or three. As the attacker can choose up to three dice, two defending dice favors the attacker;
    \item Maximum troops moved to a conquered territory: it is either the number of battle-surviving troops minus one or the number of all troops in the attacking territory minus one. The latter favors that a recently occupied territory becomes the starting point of a new attack;
    \item Bonus troop factor: at the beginning of each turn, each player receives troops equal to the number of owned territories divided by this factor, which can be 1, 2, 3 or 4. The player receives 3 troops if the division falls below three;
    \item Map: the map is represented by a graph. Every possible connected graph is a valid game map, hence the possible maps are infinite.
\end{itemize}

%Finally, the map can vary in several ways. During the crossover process, parts of the parents' maps are merged, creating a map with characteristics from both. This process can lead to changes in the number of territories, especially when one parent has a significantly different number of territories compared to the other. In the mutation phase, there is a possibility to remove, add, or move a territory, as well as exchange the troop value between two continents or add 1 troop to that value. The resulting map from this process needs to follow certain rules, being planar to represent a map and connected so that all territories are reachable. The map has the most significant influence on game duration. The larger the map, the longer it takes to play. It's important to note that there is no limit to the map's growth. The graph can expand without any size restrictions. Therefore, the search space for optimization is infinite.

\subsection{Genetic Algorithm}
\label{sec:ga}

\begin{figure}[!ht]
    \centering
    \includegraphics[width=1\columnwidth]{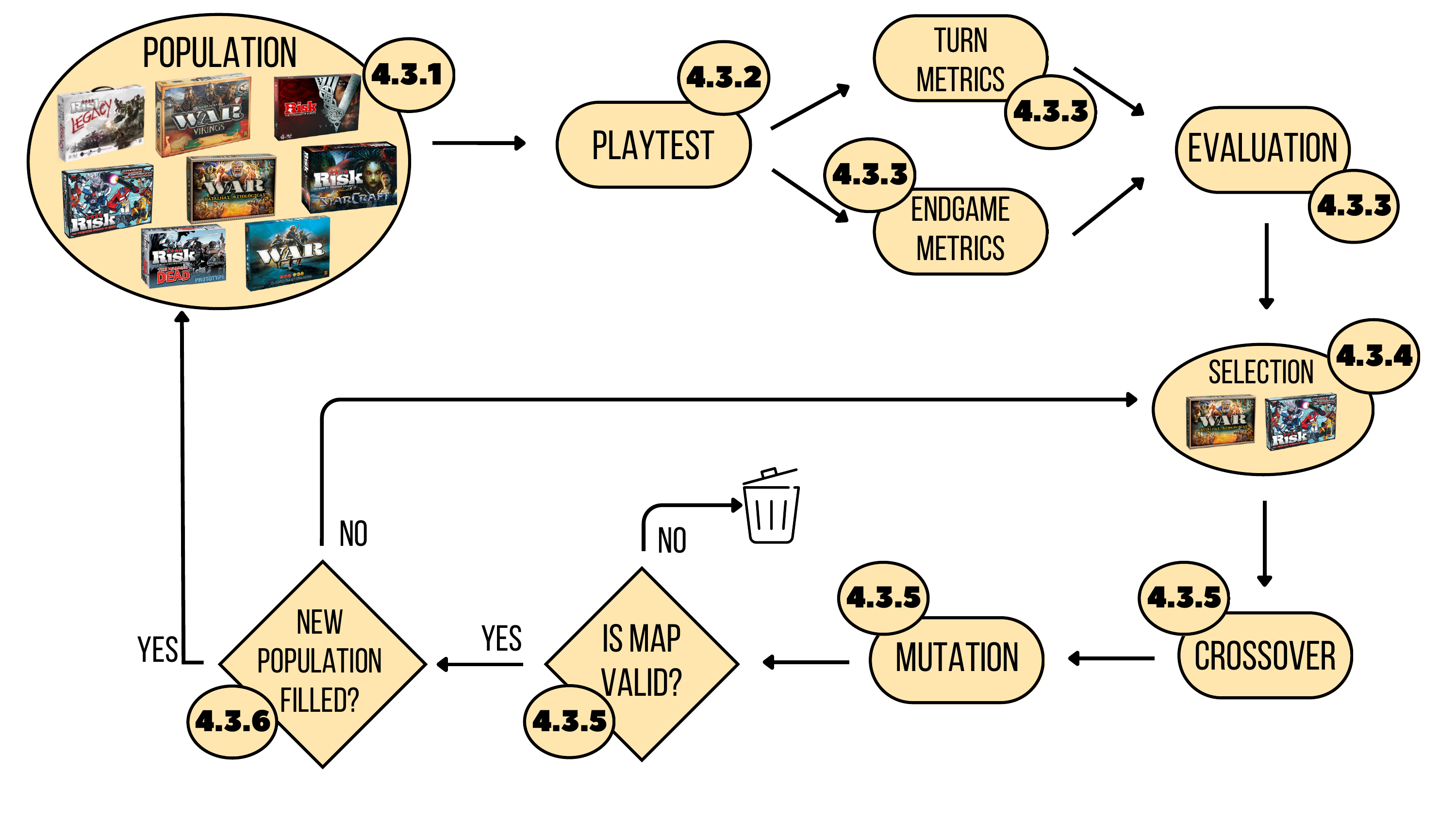}
    \caption{Genetic algorithm for generating Risk game versions. Numbers indicate the subsection describing each component. Playtests generate turn and endgame metrics, being described in the same subsection. Crossover, mutation and map validity are also described in a single subsection.}
    \label{pipeline}
\end{figure}

Our Genetic Algorithm (GA) follow the usual processes of selection, reproduction, and replacement, detailed next.
The stopping criterion on this work is the number of generations. %Although we may have a reference version with an optimal fitness according to the evaluation criteria, this version is not used to determine the algorithm's stopping point. Similarly, a parameter combination that achieves the ideal version of the evaluation metrics is not known.
%During reproduction, the newly created map graphs may have difficult-to-find or resolve errors. However, some errors are simple and are resolved before moving to the next step.
This entire process is presented in Figure \ref{pipeline}.% where the basis is a genetic algorithm. % as a minimization problem, aiming to find the individual with the lowest possible fitness. Test and evaluation steps are added due to the nature of the work. 

\subsubsection{Initial Population}

The initial population is randomly created. The parameter values for each individual are chosen from the predefined possibilities. Possible starting maps include 5 official maps and 5 artificial ones. The official maps, although balanced, are large, with 42 territories divided into 6 continents, often resulting in long games. They were taken from the original risk and the variations: StarCraft, Transformer Edition, Star Wars: Clone Wars and Halo Wars.
The 5 artificial maps are smaller, ranging from 9 to 17 territories. They aim to create more diverse individuals. %This allows for the exploration of different scenarios and brings more variety to the generated games. 

\subsubsection{Playtest}
\label{sec:playtest}
As reported by \cite{Bombardelli2022}, initial experiments have shown that Risk was challenging for basic Monte Carlo Tree Search (MCTS). It would either require too much time to perform a decent move, delaying the entire evolutionary process, or return poor decisions if allowed a short time. 
Thus, each individual of the GA is evaluated through a playtest composed of J matches between handcrafted rule-based players, described next. %resulting in poor decisions within limited time. Considering this context, a simple rule-based agent was developed by Bombardelli and will also be used in this dissertation. This agent has been updated based on revisions made in the used version of the Risk game.

During the allocation phase, the rule-based agent tries to place its troops in territories within the continent that provides the highest bonus troops. By conquering this continent early in the game, the agent gains the maximum troop bonus in the first attack phase. %If no empty continents are available, the agent chooses randomly. When there are continent bonuses, troops are placed where there are fewer units, but always on the borders of the continent that provided the bonus.
The agent performs card exchanges when it owns a territory on its cards to gain extra troops. Otherwise, it holds the cards until the maximum of 5, where an exchange is mandatory. %If the agent has 3 or 4 cards and does not possess a card with a territory it owns, the exchange is not made. However, if it has 5 cards, the exchange is mandatory according to the game rules.
When attacking, the agent looks for battles where it has a numerical advantage. %, meaning it attacks territories with fewer troops than its own, aiming to conquer as many territories as possible while having enough troops. 
If there are no such situations, the agent stops attacking. 

In the fortification phase, the agent follows the same strategy as in the allocation phase. It moves troops to less populated territories that are bordering the opponent. Non-border territories can remain with only one troop.

%With these characteristics, it is possible to play Risk sensibly without using more sophisticated strategies. Both players in the testing phase use the same agent. 
The described agent is fast, allowing for the evaluation of many games, which is necessary for the genetic algorithm.

During and at the end of a match, we calculate some metrics regarding the plays in order to evaluate the game in the next step of the pipeline. 

\subsubsection{Evaluation and Fitness}

Each game is evaluated for its quality. However, assessing how fun, interesting, or engaging a game is can be challenging, as the perception of fun and entertainment is subjective. Nevertheless, it is possible to evaluate some characteristics related to the quality of a game from the players' perspective. 
\citeauthor{browne2011evolutionary} presented 57 criteria for measuring the quality of a game, and we use 7 of these criteria, which were considered the most relevant for our purposes. 
Criteria related to rules and equipment were discarded because we generate games with the same equipment and base rules.
Also, criteria that evaluate characteristics such piece color, that are not relevant in Risk, or based on lookahead searches, which are not performed due to the game complexity (see Section \ref{sec:playtest}), were discarded.
Other criteria were either complementary or redundant to the ones we chose. For example: tendency of draws (discarded) versus completion (included, see below).

The criteria are described below, and their equations are in Appendix \ref{equations}. 

\emph{Completion} improves when a game results in victories more often than draws, regardless of the player. 

%Risk may seem never-ending, but a game will always come to an end. However, 
The \emph{duration} criteria measure how long a game tends to be, because it is neither desirable for a game to last too long as it can become tiresome and monotonous nor too short to become trivial. %, contrary to the expected qualities of the generated versions, so we calculate the duration.

The \emph{advantage} metric measures the unbalance of the game in favor of the first player. It is necessary to maintain balance in order to keep both players engaged. %, resulting in the advantage metric.

The \emph{branching factor} (number of possible moves per turn) indicates the complexity of a game, that is, how many possibilities the player needs to consider. % in order to make the best decision each turn, being the branching factor. 
%During a match, several metrics are measured per turn. The number of possible moves for a player is measured only at the beginning of each phase. 

The \emph{drama} measures the possibility, or expectation, that a player has to turn the tide of a losing game. %When a player is losing, they must have at least the hope of being able to win. 
If there is no chance of winning (low drama), the game will no longer be interesting for the player. %Since the main objective of a game is to win a match, one way to captivate the player's attention is to feed their hope of winning, being calculated by the drama.

\emph{Killer moves} significantly change a player's situation, turning them from a loser to a winner quickly. Such moves are interesting since they can dramatically change the evaluation of the players at that moment. 

\emph{Lead changes:} similar to drama, if a game does not alternate the leader, it becomes tedious. 
However, too much lead changes make the game too unpredictable or chaotic.
%Without variation in the lead, the player who starts winning will end up winning as well, making the game uninteresting for the one who is losing since they know they will never have a chance to turn the game around.
Turn metrics measure events inside a match whereas endgame metrics are used to measure characteristics related to the game termination or result. Our turn metrics are branching factor, drama, killer moves and lead changes, whereas our endgame metrics are completion, duration and advantage.

%\subsubsection{Fitness}

\textbf{Fitness function:} for fitness calculation, the approach tested was based on distance. In this method, optimal values are chosen for each evaluation criteria. The distance between the analyzed individual and an optimal individual is then calculated. The optimal values can be seen in Table \ref{criteriosotimos}.

\begin{table}[!ht]
\centering
\caption{Optimal values for the evaluation criteria.}
\label{criteriosotimos}
\begin{tabular}{r|l}
    \toprule
    Criterion             & Optimal Value \\ \midrule
    Completion            & 1 (does not end up in draws) \\           
    Duration              & 0 (each game ends in 24 turns) \\
    Advantage             & 0 (50\% wins for both players) \\
    Branching Factor      & 0.5 (avg. of 50 possible moves per turn) \\
    Drama                 & 0.5 (avg. degree of drama) \\    
    Killer Moves          & 0.5 (avg. degree of killer moves) \\
    Lead Change           & 0.5 (avg. degree of lead change) \\
    \bottomrule
\end{tabular}
\end{table}

The fitness calculation is performed as shown in Equation \ref{calculoFitness}, where  $v_i$ is the i-th criterion of the individual $v$, \textit{I} is the set of criteria, and $o_i$ is the optimal value of criterion $i$. 
%calculated by summing the absolute value of the subtraction of criterion \textit{i} of version \textit{c\textsubscript{vi}} from the optimal value of the same criterion \textit{ot\textsubscript{i}}, where \textit{I} is a list containing all the criteria.

\begin{equation}
\label{calculoFitness}
    fitness_v = \sum_{i=1}^I {|v_i - o_i|}
\end{equation}

We highlight that our fitness imposes a minimization problem, i.e., the lower the fitness the better. This is a mismatch with the evolutionary metaphor of ``survival of the fittest'', but we maintain our formulation to keep Eq. \ref{calculoFitness} simple and meaningful (the closer to the desired values, the better) rather than changing it towards a fitness-maximizing equation.

\subsubsection{Selection}
The GA uses tournament selection. Each tournament has k participants randomly drawn from the population, and the two with the best fitness become the parents of two new individuals.
Moreover, elitism is applied, where the best individual from the previous generation is copied to the new population. %This ensures that all the progress made so far is not lost or forgotten.

\subsubsection{Crossover and Mutation}

After selecting two individuals from the population, we use a crossover function to generate a new individual inheriting characteristic from both of the selected individuals. The crossover is done by selecting some continents from the first and second parents and combining them in a new map, creating new connections between these continents' territories. Now, on the new map, these continents will inherit the same territory structure and bonus units that the parent continents have.

The crossover and mutation operations are divided into two parts: attributes and the map. The attributes are recombined using a random mask that always splits the individual in half. Half of one parent is combined with the complementary half of the other parent, generating one child, and the reverse combination generates the second child. During attribute mutation, if the mutation rate is reached, the new parameter value is flipped or chosen from the possibilities, if applicable. Each mutation occurs independently, meaning that one mutation does not imply that other parameters will mutate as well. Multiple changes can occur in the same individual.

Map operations are more complex. During crossover, some continents from both parents are chosen to be combined. They are brought with their original connections, and additional connections are created to maintain coherence when merging. %All original territories and connections are preserved during this merging. 
During mutation, several operations can be performed, including creating a new connection, removing a connection, moving a territory from one continent to another, swapping the bonus value between two continents, or modifying the continent bonus value by 1 unit. 

The map resulting from this process needs to follow some rules. It must be planar, to represent a map, and connected, so that all territories are reached. We did this verification using the Networkx library. Planarity is checked by the Left-Right Planarity Test \cite{boyer2003stop} and connectivity is checked with a depth-first search algorithm. % and the \textit{is\_planar}\footnote{https://networkx.org/documentation/stable/reference/algorithms/generated/ \\ networkx.algorithms.planarity.is\_planar.html} and \textit{is\_strongly\_connected}\footnote{https://networkx.org/documentation/stable/reference/algorithms/generated/ \\ networkx.algorithms.components.is\_strongly\_connected.html}.

\subsubsection{Replacement}

After each iteration, the entire population used to generate the next one is replaced. Thus, the new population consists of the best individual from the previous population (elitism), along with the offspring produced by that population.

\section{Experiments}

The first step of the experiments was to find the best values for the hyperparameters of the genetic algorithm. After that, we show how maps evolved and their main characteristics in the final generations. Then we show the results of the evolution of the remaining parameters.
We remark that, in our formulation, we aim for as low fitness as possible, as it indicates that the individual has metrics approaching the desired values.

%Some gameplay tests and tests with maps of the original games were also performed. In this stage, the goal is to understand how the pipeline behaves with parameter sets related to real games and how far it can go by varying these games.

\subsection{Hyperparameters}
\label{sec:hyperparams}

During this initial testing phase, different values of 4 hyperparameters were analyzed. 100 matches are played for each generated game, with a maximum limit of 48 turns and a maximum time of 60 seconds per match. These hyperparameters were chosen to consider the time for game testing, aiming to find the best combination for future tests.

The parameters and tested values are as follows:

\begin{itemize}
    \item Number of generations: 10, 30, 50, 70, 90, 110, 130, 150, 170, 190. %Starting from 10 generations, which is the minimum value chosen to generate reliable results, the values increased in increments of 20 generations.
    \item Offspring size: 5, 10, 15, 20, 25, 30, 35, 40, 45, 50. %Starting with 5, the values increased in increments of 5 until reaching 50 individuals per generation.
    \item Tournament size: the number of participants in the selection tournament can vary from 2 to half the number of individuals in the population, rounded down, in increments of 2. For example, if the population size is 30, the tournament participants range from 2 to 14.
    \item Mutation rate: 0.1, 0.2, 0.4, 0.6, 0.8. The rate of 0.1 was added, considering the parameter used by \citeauthor{browne2008automatic}.
\end{itemize}

All possible combinations of these 4 hyperparameters were tested to evaluate and find the best option, i.e., the one that resulted in a fitness value closest to the optimal value, according to the distance-based calculation. In total, there were 3,250 distinct executions. Table \ref{top10table} shows the fitness of the top 10 executions, identified as 1 to 10, with the hyperparameters. It is important to note that these values were obtained at the end of the last generation of each execution.

\begin{table}[!ht]
\begin{center}
\caption{Parameters of each run generating the top 10 fitness}
\label{top10table}
\begin{tabular}{c|c|c|c|c|c}
    \toprule
    Run & Fitness & \makecell{Num. of \\ Gen.} & \makecell{Off. \\ Size} & \makecell{Tourn \\ Size} & \makecell{Mutation \\ Rate} \\ \midrule
    1  & 0.779 & 10  & 50 & 22 & 0.6  \\
    2  & 0.798 & 150 & 30 & 12 & 0.1  \\
    3  & 0.800 & 50  & 50 & 16 & 0.6  \\
    4  & 0.823 & 150 & 30 & 8  & 0.6  \\
    5  & 0.847 & 150 & 20 & 6  & 0.8  \\
    \bottomrule
\end{tabular}
\end{center}
\end{table}

\begin{figure}[!ht]
    \centering
    \begin{subfigure}[t]{\linewidth}
        \centering
        \includegraphics[width=0.9\columnwidth]{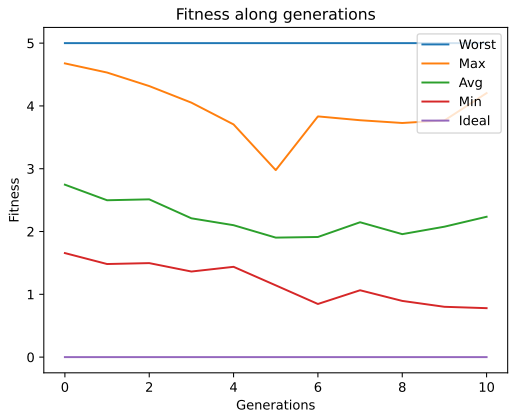}
        \caption{Configuration 1 (best): 10 generations, 50 individuals in the offspring, 22 individuals in the tournament, 0.6 mutation rate}
        \label{10gen50off22tour06mut}
    \end{subfigure}
    \hfill
    \begin{subfigure}[t]{\linewidth}
        \centering 
        \includegraphics[width=0.9\columnwidth]{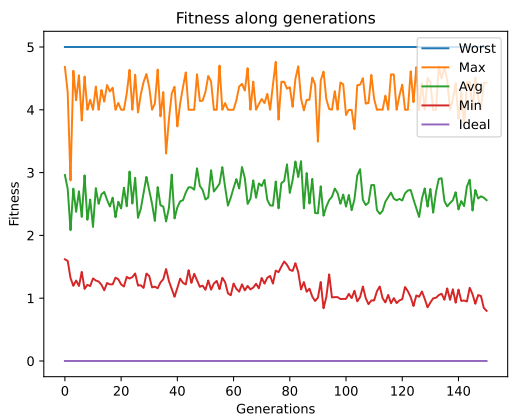}
        \caption{Configuration 2 (2nd best): 150 generations, 30 individuals in offspring, 12 individuals in the tournament, 0.1 mutation rate}
        \label{150gen30off12tour01mut}
    \end{subfigure}
    \caption{Fitness throughout the execution that resulted in the best fitness and the second-best fitness at the end of the process.}
    \label{figcomsubfig}
\end{figure}

Analyzing the parameters that led to a lower fitness value, it is interesting to note that a lower number of generations was found. In this context, it is expected that a low number of generations is not sufficient to evolve individuals to their maximum. To understand the situation, Figure \ref{figcomsubfig} shows the fitness over generations for the best (a) and second-best (b) hyperparameter configurations (configurations 1 and 2 in Table \ref{top10table}). It can be observed that in the second-best execution, there was significant variation in the lowest fitness value of each generation. In the best execution, since the number of generations is much lower, this oscillation was smoother.

When elitism is applied in the replacement of a population, the best individual from the previous population is retained in the composition of the new population. When this happens, the individual is reevaluated. The evaluation is stochastic due to randomness in dice rolls. This can cause the previously best individual to not have as good an evaluation as before. Analyzing Figure \ref{10gen50off22tour06mut}, the minimum fitness value was quickly found, and since there was no time for the various changes seen in Figure \ref{150gen30off12tour01mut} to occur, it remained stable.

Therefore, a high number of generations can lead to a low fitness value, and that is what is expected. However, with a high number of generations, there is a possibility that a low value will appear and be lost in a new evaluation.

\begin{figure}[!ht]
    \centering
    \includegraphics[width=0.8\columnwidth]{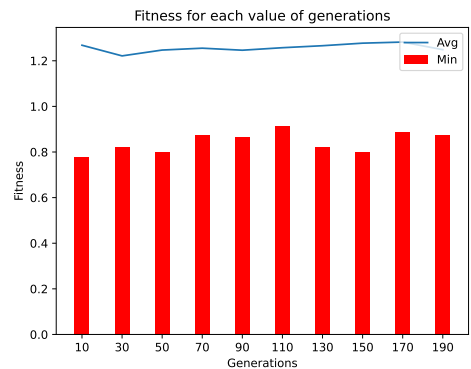}
    \caption{Final fitness versus number of generations.}
    \label{fitnessalonggen}
\end{figure}

Continuing with the analysis of hyperparameters, we can extract the best values for each of them independently and also understand how each characteristic influences the fitness outcome. When analyzing the number of generations (Figure \ref{fitnessalonggen}), it was not possible to identify a clear pattern. The values shown in the graphs are the average of each set of runs and the lowest value found in the last generation. The minimum fitness starts low, increases, and then decreases with 150 generations of execution, while the average value remains relatively stable. This suggests that the number of generations evaluated alone does not have a significant influence on the fitness outcome, as no significant pattern was identified.

\begin{figure}[!ht]
    \centering
    \includegraphics[width=0.8\columnwidth]{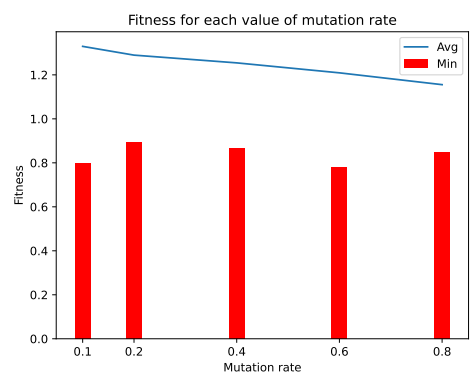}
    \caption{Final fitness versus mutation rates.}
    \label{fitnessalongmut}
\end{figure}

Considering the mutation rate (Figure \ref{fitnessalongmut}) we observe an effect on both the minimum and average values. The higher the mutation rate, the lower the average fitness. This indicates that this characteristic may not have a drastic impact on the minimum value but contributes to reducing the average value. 

When analyzing the number of individuals in the tournament and in the offspring, we observe a pattern similar to the offspring size. There is variation in the minimum fitness values and a decrease in the average as the tournament size increases. We conclude that this hyperparameter is not a measure to find the minimum fitness, but rather to decrease the values in general. However, the best fitness values were obtained with varied tournament sizes, although smaller values prevail.

In summary, no hyperparameter has complete responsibility for the final fitness values. What occurs is a decrease in the average values, improving the population as a whole. However, these values do not come very close to the ideal or minimum value. Nevertheless, the runs used in the next tests will be selected among the top 10 shown in Table \ref{top10table}.

\subsection{Map Evolution}
\label{sec:exp-map}
This section is dedicated to evaluating one of the most important aspects of the game: the map, since it can influence all evaluation metrics. The branching factor and duration are the most affected, as the possible movements consider the number of territories on the map, and the more territories, the longer it takes to conquer the entire map. In this section, the isomorphism of maps and their viability are evaluated. Isomorphism was analyzed using the VF2 algorithm \cite{Cordella2001isomorphism} implemented in the Networkx library.

The first step is to identify maps with isomorphic graphs to verify if there were populations dominated by one map. Two runs exhibited this characteristic, meaning that all maps in the last generation of both runs were isomorphic to an initial map. Both runs had 10 generations and a 10\% mutation rate. These two factors may have strongly contributed to the propagation of only one map. From the initial population, only 2 individuals carried this map. In the final population, all individuals had maps that were isomorphic to Map 9.

%A total of 65 combinations of isomorphic maps were found, totaling 30 individuals in the final population of one run and 35 individuals in the final population of another run. However, these are only two parameter configurations out of a total of 3250 evaluated combinations. This does not make the evolution process considered poor, since this occurrence was not recurrent.

The main strength of Map 9 is the small number of territories compared to actual Risk maps. With 13 territories, the game time is significantly reduced, as is the branching factor. Although it does not drastically affect the other criteria, the decrease in the number of possible movements is something that not all maps can offer.

On average, the best individuals from all test runs had a minimum of 2 territories, a maximum of 24 territories, and an average of 3.39 territories. Even the individual with the highest number of territories has fewer territories than real Risk maps, which have 42 territories.

In another analysis, a comparison of maps from the fastest run and the longest run was conducted to understand the effects of extreme hyperparameter values. The hyperparameters for the fastest run are 10 generations, 5 individuals in the population, 2 as the tournament size, and 0.1 mutation rate. The hyperparameters for the longest run are 190 generations, 50 individuals in the population, 24 as the tournament size, and 0.8 mutation rate. 

\begin{figure}[!ht]
    \centering
    \begin{subfigure}[t]{\linewidth}
        \centering
        \includegraphics[width=0.6\columnwidth]{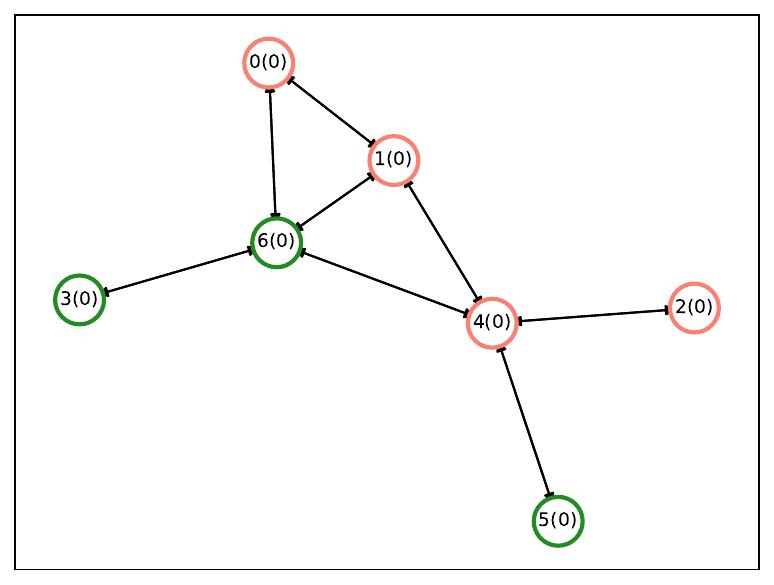}
        \caption{Map representation of the best individual of the fast run.}
        \label{mapa49}
    \end{subfigure}
    \hfill
    \begin{subfigure}[t]{\linewidth}
        \centering 
        \includegraphics[width=0.6\columnwidth]{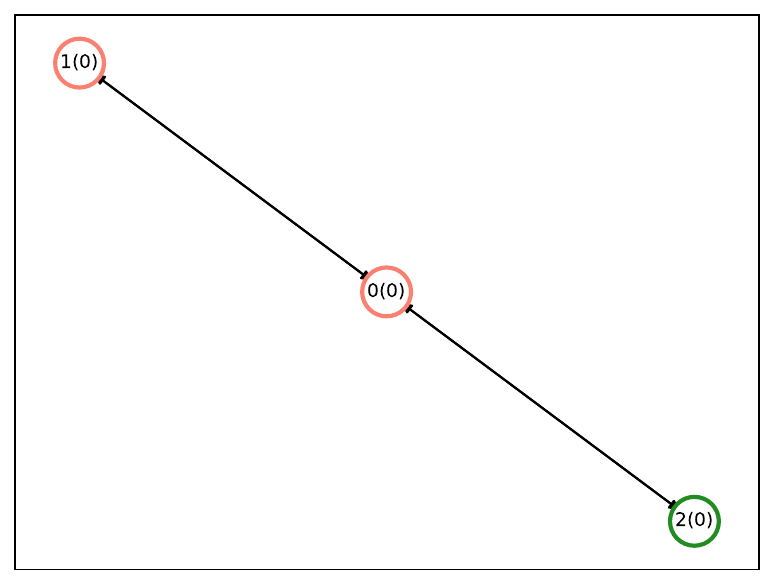}
        \caption{Map representation of the best individual of the long run.}
        \label{mapa13655}
    \end{subfigure}
    \caption{Map representation of the best individual of the fast and the long run.}
    \label{figcomsubfigmapas}
\end{figure}

The map of the best individual generated by the fastest run can be seen in Figure \ref{mapa49}, and the map from the longest run is shown in Figure \ref{mapa13655}. Both maps have a significantly reduced number of territories, with the fast run map having 7 territories and the long run map having 2 territories, very close to the average.

Looking at the criteria shown in Table \ref{criteriasimpleandexhaustive}, we can see why the best individuals had these maps. The advantage of both maps is very close to the ideal value, with the fast run achieving this ideal value, and the branching factor also remained close to the ideal value. Additionally, the duration was also reduced in both runs.

\begin{table}[!ht]
\begin{center}
\caption{Scoring of the fastest and longest runs in each criteria and their respective distances from ideal values.}
\label{criteriasimpleandexhaustive}
\begin{tabular}{c|c|c|c|c|c}
    \toprule
    Criteria & Ideal & Fast & \makecell{Fast \\ Dist.} & Long  & \makecell{Long \\ Dist.} \\ \midrule
    Completion               & 1    & 0.98 & 0.02        & 1    & 0    \\
    Duration                 & 0    & 0.41 & 0.41        & 0.69 & 0.69 \\
    Advantage                & 0    & 0    & 0           & 0.02 & 0.02 \\
    Branching F.             & 0.5  & 0.76 & 0.26        & 0.50 & 0    \\
    Drama                    & 0.5  & 0.30 & 0.2         & 0.42 & 0.08 \\ 
    Killer Moves             & 0.5  & 0.78 & 0.28        & 0.56 & 0.06 \\
    Lead Change              & 0.5  & 0.25 & 0.25        & 0.42 & 0.08 \\ 
    \textbf{\textit{Fitness}} & \textbf{0} &      & \textbf{1.42} &      & \textbf{0.93} \\
     \bottomrule
\end{tabular}
\end{center}
\end{table}

The values considered ideal for the criteria have a limitation. The duration and branching factor criteria favor the emergence of small maps, such as the one in Figure \ref{mapa49}, as the number of territories strongly influences these criteria. Therefore, since the fitness function requires low values for these criteria, the number of territories needs to be small. %In maps like the one in Figure 5.9, the victory of the first player is almost certain, which is undesirable even if it satisfies the defined fitness function.

Among all 3250 executions, 3236 (99.56\%) have their best individual with up to 10 territories. This shows a strong predominance of small maps, and it is possible to conclude that in order to achieve ideal criterion values, the generated maps need to be small.

\subsection{Game Evolution}
\label{sec:exp-game}
After understanding the evolution of maps, we need to also examine the remaining parameters. Although they have a smaller set of combinations, they are equally important as the map in fulfilling the objectives of this work.

Overall, the parameters do not have a predominant influence on the final generation. The two values for the initial territory distribution mode parameter were evenly divided:

\begin{itemize}
    \item Randomly distributed territories: 1900
    \item Players choose territories: 1350
\end{itemize}

These values suggest that this parameter has little influence on the game performance. When players choose their territories, there is a high chance that both will start the game with the advantage of owning an entire continent. When the distribution is random, neither player has this advantage. %Thus, in both modes of territory distribution, the game is equally balanced but with a tendency towards random distribution.

Looking at the defensive dice quantity parameter, there is some difference in the number of games with each value, but it is not very significant:

\begin{itemize}
    \item 2 dice: 1674
    \item 3 dice: 1576
\end{itemize}

When only 2 defensive dice are used, there is an advantage for the attacker, as the player has one more chance to roll a higher number than the opponent. With 3 dice, this advantage decreases, as both players have an equal number of dice. %Therefore, we can see a slight tendency to favor the attacker.

In the parameter of the number of troops moved when conquering a new territory, a larger difference can be observed:

\begin{itemize}
    \item Minimum quantity (number of attacking dice - 1): 2629
    \item Maximum quantity (number of troops in the attacking territory - 1): 621
\end{itemize}

When the minimum quantity is transferred to the new territory, the number of possible attacks from this conquered territory is small and likely limits the player's consecutive actions. With the maximum quantity, the game can become longer. %This parameter considerably impacts the game duration, as it indicates how far the player can go.

Lastly, the factor that calculates the number of troops received at the beginning of each turn obtained more uniform results:

\begin{itemize}
    \item Factor = 1: 773
    \item Factor = 2: 807
    \item Factor = 3: 845
    \item Factor = 4: 825
\end{itemize}

Considering the minimum number of troops that must be received by the player at the start of their turn, which is 3, they need to have 12 territories to obtain that amount. As mentioned earlier in Section 5.2, the average number of territories in the maps of the individuals in the last generation of the test runs was 3.39. Therefore, it is impossible to obtain more than 3 troops, rendering the factor irrelevant.

%Based on these results, we can conclude that the best individual would have randomly distributed territories, 3 dice for the defender, the minimum quantity of troops moved when conquering a territory, and a factor of 3. The individual with the best fitness has exactly these parameters.

\subsection{Gameplay, Generated Versions and Real Versions}
\label{sec:exp-gameplay}
Testing the gameplay of the best generated games requires many responsible human participants with available time. Therefore, due to time limitations, the author was the only one who conducted real tests.

The tested games are playable and consistent with the ideal criterion values. However, it is only through tests with human players that it will be possible to evaluate if the generated games are fun, challenging, and engaging enough.

Considering that the map size is the factor that can vary the most in the evolution process, games from the initial generations have larger maps than those generated in the end. However, this criterion alone is not sufficient to determine if a game is good or bad, as there may be a combination of other parameters that make a game good on a map considered bad.

The evaluation of Risk versions can be compared with the generated games. The parameters for Risk include free initial territory choice, a factor of 3 in the calculation of received troops, 2 defense dice, and the maximum number of troops moved when conquering a new territory. It is also important to note that the maximum number of turns played before declaring the game unfinished is 24, with 12 turns for each player. The values of the criteria for Risk, along with the ideal criteria and the best-generated game, are listed in Table \ref{realcriteria}.

\begin{table}[!ht]
\begin{center}
\caption{Values of criteria calculated for the original parameters of Risk}
\label{realcriteria}
\begin{tabular}{c|c|c|c}
    \toprule
    Criteria              & Risk  & Ideal  & Best Gen. \\ \midrule
    Completion            & 0.063 & 1      & 1 \\
    Drama                 & 0.890 & 0.5    & 0.427 \\
    Duration              & 0.994 & 0      & 0.637 \\           
    Branching F.          & 1     & 0.5    & 0.514 \\
    Killer Moves          & 0.273 & 0.5    & 0.475 \\
    Lead Change           & 0.009 & 0.5    & 0.489 \\
    Advantage             & 0.96  & 0      & 0.02 \\ 
    \textbf{\textit{Fitness}} & \textbf{4.498} & \textbf{0} & \textbf{0.779} \\  \bottomrule
\end{tabular}
\end{center}
\end{table}

The branching factor for both games was 1, indicating that there are many possible moves in each turn. When calculating the branching factor, the possibilities are truncated at 100. So, if the branching factor is equal to 1, it means that the player had an average of 100 possible moves in the test round. Additionally, the advantage for Risk games was high, 0.96, indicating that the first player has a higher chance of winning than the second. The completeness for Risk was extremely low, reaching the limit of turns imposed, which is 24. 

The duration of Risk has a very high value, indicating that the games last much longer than ideal. Also, Risk has a balanced advantage. The other metrics help stabilize the evaluation and represent a significant part of the fun of a game. The original parameters of Risk result in different versions of the same game with distinct metrics.

It is important to note that the parameters are applied to a simplified implementation of the game and are played with a rule-based agent. The original parameters in the original rules are suitable for human players, given the commercial success of the games.

\subsection{Analysis}
\label{sec:analysis}
In the first experiment, different hyperparameters were tested to identify combinations that result in individuals with higher fitness. The results are summarized in Table \ref{top10table}. Analysis of Figures \ref{fitnessalonggen} and \ref{fitnessalongmut} showed that increasing the values of the hyperparameters only decreased the average fitness over generations.

It was observed that the evolutionary process tends to generate smaller maps, with 99.56\% of the best individuals having 10 territories or fewer. This pattern likely occurs due to the attempt to reach the desired values of branching factor and duration metrics.

Furthermore, analysis of the other parameters revealed that certain values favor the generation of better games. The random distribution of territories, the use of 2 dice for the defender, the minimum number of troops moved in conquering a territory, and a factor of 3 were the most common parameters among the best individuals, including the individual with the best fitness.

%Finally, tests were conducted with the parameters of the Risk game in the simplified version of Risk presented in the study. Both games achieved high fitness values compared to the best version generated by the genetic algorithm.

Considering the experiments and results presented, we can conclude that the pipeline is capable of generating new variations of the Risk game. All generated variations are playable; however, those that underwent more generations resulted in small maps, with an average of 3.39 territories. Those result in nearly trivial games.

A limitation of this work is that the evolution process heavily depends on the agent used for the game. It is necessary to implement a skilled agent to allow the generation of versions with more strategic depth. 
The generated games are evaluated based on the players' actions, and if they play in a basic manner, the games also tend to be basic. 
The rule-based player is fast but simple. On the other hand, AI algorithms that perform well in board games are not efficient for tabletop games.

The use of the genetic algorithm brings a new approach to the experiments that were previously conducted with genetic programming. With the genetic algorithm, the parameter exploration is more aggressive and can provide greater variability to the generated individuals.

\section{Conclusion}
\label{sec:conclusion}

This work expanded the evolutionary game design process to include tabletop games, being assessed in Risk, a territorial-domination military strategy game. The game is represented by basic, immutable rules, but its behavior is highly customized by the exposed set of parameters. The genetic algorithm used aims to find the best possible combination of these parameters. The fitness function is calculated by summing the distances of quality metrics to their desired values. The metrics are assessed via playtests with a rule-based agent.

During the execution of this work, various approaches were explored for the game Risk, which had been little explored until then. From expanding a rule-based agent to creating a new version of the game, different methodologies were tested in different areas. It was possible to explore different ways to analyze game quality and understand how they influence the evolution parameters.

% understand graph crossovers and their main difficulties in integration, as well as

For future work, it is important to address the limitations identified in this study. The first one is the fitness instability due to the randomness of the game. Increasing the number of matches to calculate the metrics can make the values more stable. Furthermore, a more in-depth study of the metrics that make up the fitness and their target values can improve the quality of the generated games, since in our tests, very small maps that induced nearly trivial games had good fitness. %Procedural map generation can also increase the diversity of the initial population, resulting in more varied maps.
Developing a more sophisticated agent is another area of utmost importance. The challenge is to create an agent that efficiently handles the complexity of the game, as a simple fitness evaluation requires hundreds of matches. Exploring extensions of techniques such as MCTS \cite{browne2012survey} or RMEA \cite{islam2007}, used in digital games, may be promising.

Additionally, tests with human players can provide different perspectives and insights into the game and its versions. In the future, the continuation of this work will require test groups to find better versions and also to define ideal values for evaluation metrics.
Finally, adapting this work to a wide range of games is an interesting project. Challenges such as including cards and pieces with different movements can be explored. An even greater challenge would be the establishment of a game description language for board games, enabling the creation of completely new games.

%%%%%%%%%%%%% Acknowledgements %%%%%%%%%%%%%
%\footnotesize
\section*{Acknowledgements}

This study was financed in part by the Coordenação de Aperfeiçoamento de Pessoal de Nível Superior - Brasil (CAPES) - Finance Code 001.

%%%%%%%%%%%%%%   Bibliography   %%%%%%%%%%%%%%
\normalsize
\bibliography{references}

\begin{thebibliography}{29}
\providecommand{\natexlab}[1]{#1}
\providecommand{\url}[1]{\texttt{#1}}
\expandafter\ifx\csname urlstyle\endcsname\relax
  \providecommand{\doi}[1]{doi: #1}\else
  \providecommand{\doi}{doi: \begingroup \urlstyle{rm}\Url}\fi

\bibitem[Browne(2011)]{browne2011evolutionary}
Cameron Browne.
\newblock \emph{Evolutionary Game Design}.
\newblock SpringerBriefs in Computer Science. Springer London, London, UK, 2011.
\newblock ISBN 9781447121794.
\newblock URL \url{https://doi.org/10.1007/978-1-4471-2179-4}.

\bibitem[Genesereth et~al.(2005)Genesereth, Love, and Pell]{Genesereth2005ggp}
Michael Genesereth, Nathaniel Love, and Barney Pell.
\newblock {General game playing: Overview of the AAAI competition}.
\newblock \emph{AI magazine}, 26\penalty0 (2):\penalty0 62--62, 2005.

\bibitem[Browne et~al.(2012)Browne, Powley, Whitehouse, Lucas, Cowling, Rohlfshagen, Tavener, Perez, Samothrakis, and Colton]{browne2012survey}
Cameron~B Browne, Edward Powley, Daniel Whitehouse, Simon~M Lucas, Peter~I Cowling, Philipp Rohlfshagen, Stephen Tavener, Diego Perez, Spyridon Samothrakis, and Simon Colton.
\newblock A survey of monte carlo tree search methods.
\newblock \emph{IEEE Transactions on Computational Intelligence and AI in games}, 4\penalty0 (1):\penalty0 1--43, 2012.

\bibitem[Browne(2018)]{Browne2018ludii}
Cameron Browne.
\newblock {Modern Techniques for Ancient Games}.
\newblock In \emph{IEEE Conference on Computational Intelligence and Games}, pages 490--497, Maastricht, 2018. IEEE Press.

\bibitem[Hendrikx et~al.(2013)Hendrikx, Meijer, Van Der~Velden, and Iosup]{hendrikx2013}
Mark Hendrikx, Sebastiaan Meijer, Joeri Van Der~Velden, and Alexandru Iosup.
\newblock Procedural content generation for games: A survey.
\newblock \emph{ACM Transactions on Multimedia Computing, Communications, and Applications}, 9\penalty0 (1), feb 2013.
\newblock ISSN 1551-6857.
\newblock \doi{10.1145/2422956.2422957}.
\newblock URL \url{https://doi.org/10.1145/2422956.2422957}.

\bibitem[Togelius et~al.(2011)Togelius, Yannakakis, Stanley, and Browne]{togelius2011}
Julian Togelius, Georgios~N. Yannakakis, Kenneth~O. Stanley, and Cameron Browne.
\newblock Search-based procedural content generation: A taxonomy and survey.
\newblock \emph{IEEE Transactions on Computational Intelligence and AI in Games}, 3\penalty0 (3):\penalty0 172--186, 2011.
\newblock \doi{10.1109/TCIAIG.2011.2148116}.

\bibitem[Summerville and Mateas(2016)]{summerville2016mystical}
Adam Summerville and Michael Mateas.
\newblock Mystical tutor: A magic: The gathering design assistant via denoising sequence-to-sequence learning.
\newblock \emph{Proceedings of the AAAI Conference on Artificial Intelligence and Interactive Digital Entertainment}, 12\penalty0 (1):\penalty0 86--92, 2016.

\bibitem[Uriarte and Onta{\~{n}}{\'{o}}n(2013)]{uriarte2013}
Alberto Uriarte and Santiago Onta{\~{n}}{\'{o}}n.
\newblock {PSMAGE:} balanced map generation for starcraft.
\newblock In \emph{2013 {IEEE} Conference on Computational Inteligence in Games (CIG)}, pages 1--8, Niagara Falls, ON, Canada, 2013. {IEEE}.
\newblock \doi{10.1109/CIG.2013.6633644}.
\newblock URL \url{https://doi.org/10.1109/CIG.2013.6633644}.

\bibitem[Brand and Kroon(2014)]{Brand2014}
Dirk Brand and Steve Kroon.
\newblock Sample evaluation for action selection in {M}onte {C}arlo tree search.
\newblock In \emph{Proceedings of the Southern African Institute for Computer Scientist and Information Technologists Annual Conference 2014 on SAICSIT 2014 Empowered by Technology}, pages 314--322, Centurion, South Africa, 2014. Association for Computing Machinery.
\newblock ISBN 9781450332460.
\newblock \doi{10.1145/2664591.2664612}.
\newblock URL \url{https://dl.acm.org/doi/10.1145/2664591.2664612}.

\bibitem[Olsson(2005)]{Olsson2005}
Fredrik Olsson.
\newblock A multi-agent system for playing the board game {R}isk.
\newblock Master's thesis, Blekinge Institute of Technology, 2005.
\newblock URL \url{https://www.diva-portal.org/smash/get/diva2:831093/FULLTEXT01.pdf}.

\bibitem[Carr(2020)]{Carr2020}
Jamie Carr.
\newblock Using graph convolutional networks and {TD}($\lambda$) to play the game of risk, 2020.
\newblock URL \url{https://arxiv.org/abs/2009.06355}.

\bibitem[Heredia and Cazenave(2021)]{Heredia2021}
Lucas~Gnecco Heredia and Tristan Cazenave.
\newblock Expert iteration for {R}isk.
\newblock In \emph{Advances in Computer Games: 17th International Conference, ACG 2021}, pages 27--37, Virtual Event, 2021. Springer-Verlag.
\newblock ISBN 978-3-031-11487-8.
\newblock \doi{10.1007/978-3-031-11488-5_3}.
\newblock URL \url{https://doi.org/10.1007/978-3-031-11488-5_3}.

\bibitem[Wolf(2005)]{Wolf2005}
Michael Wolf.
\newblock \emph{An Intelligent Artificial Player for the Game of {R}isk}.
\newblock PhD thesis, Darmstadt University of Technology, 2005.

\bibitem[Harju(2012)]{harju2012probabilities}
Manu Harju.
\newblock On probabilities of {R}isk type board game combats, 2012.
\newblock URL \url{https://arxiv.org/abs/1204.4082v1}.

\bibitem[Marks(1998)]{marks1998using}
Michael~P. Marks.
\newblock Using the game of {R}isk to teach international relations.
\newblock \emph{International Studies Notes}, 23\penalty0 (1):\penalty0 11--18, 1998.
\newblock ISSN 00947768.
\newblock URL \url{http://www.jstor.org/stable/44235311}.

\bibitem[Snodgrass and Onta{\~n}{\'o}n(2017)]{snodgrass2017}
Sam Snodgrass and Santiago Onta{\~n}{\'o}n.
\newblock Learning to generate video game maps using markov models.
\newblock \emph{IEEE Transactions on Computational Intelligence and AI in Games}, 9\penalty0 (4):\penalty0 410--422, 2017.
\newblock \doi{10.1109/TCIAIG.2016.2623560}.

\bibitem[Chen and Guy(2020)]{chen2020chaos}
Tiannan Chen and Stephen Guy.
\newblock Chaos cards: Creating novel digital card games through grammatical content generation and meta-based card evaluation.
\newblock \emph{Proceedings of the AAAI Conference on Artificial Intelligence and Interactive Digital Entertainment}, 16\penalty0 (1):\penalty0 196--202, 2020.
\newblock \doi{https://doi.org/10.1609/aiide.v16i1.7430}.
\newblock URL \url{https://ojs.aaai.org/index.php/AIIDE/article/view/7430}.

\bibitem[Togelius and Schmidhuber(2008)]{Togelius2008}
Julian Togelius and Jurgen Schmidhuber.
\newblock An experiment in automatic game design.
\newblock In \emph{2008 IEEE Symposium On Computational Intelligence and Games, CIG 2008}, pages 111--118, Perth, Australia, 2008. {IEEE}.
\newblock ISBN 9781424429745.
\newblock \doi{10.1109/CIG.2008.5035629}.

\bibitem[Hom and Marks(2007)]{Hom2007}
Vincent Hom and Joe Marks.
\newblock Automatic design of balanced board games.
\newblock \emph{Proceedings of the AAAI Conference on Artificial Intelligence and Interactive Digital Entertainment}, 3:\penalty0 25--30, 2007.
\newblock ISSN 2334-0924.
\newblock \doi{10.1609/AIIDE.V3I1.18777}.
\newblock URL \url{https://ojs.aaai.org/index.php/AIIDE/article/view/18777}.

\bibitem[Wiklund et~al.(2015)Wiklund, Rudenmalm, Norberg, Westin, and Mozelius]{wiklund2015evaluating}
Mats Wiklund, William Rudenmalm, Lena Norberg, Thomas Westin, and Peter Mozelius.
\newblock Evaluating educational games using facial expression recognition software: measurement of gaming emotion.
\newblock In \emph{Proceedings of the European Conference on Games Based Learning}, pages 605--612, Steinkjer, Norway, 2015. Academic Conferences and Publishing International Limited.

\bibitem[Ferrari and Assun{\c c}{\~a}o(2022)]{ferrari2022towards}
René Ferrari and Joaquim Assun{\c c}{\~a}o.
\newblock Towards playing risk with a hybrid {M}onte {C}arlo based agent.
\newblock In \emph{Anais Estendidos do XXI Simpósio Brasileiro de Jogos e Entretenimento Digital}, pages 301--306, Porto Alegre, RS, Brasil, 2022. SBC.
\newblock \doi{10.5753/sbgames_estendido.2022.225471}.
\newblock URL \url{https://sol.sbc.org.br/index.php/sbgames_estendido/article/view/23664}.

\bibitem[Gibson et~al.(2010)Gibson, Desai, and Zhao]{gibson2010automated}
Richard Gibson, Neesha Desai, and Richard Zhao.
\newblock An automated technique for drafting territories in the board game {R}isk.
\newblock \emph{Proceedings of the AAAI Conference on Artificial Intelligence and Interactive Digital Entertainment}, 6\penalty0 (1):\penalty0 15--20, 2010.

\bibitem[Gaina et~al.(2020)Gaina, Balla, Dockhorn, Montoliu, and Perez-Liebana]{gaina2020tag}
Raluca~D. Gaina, Martin Balla, Alexander Dockhorn, Raul Montoliu, and Diego Perez-Liebana.
\newblock {TAG: A Tabletop Games Framework}.
\newblock In \emph{{Experimental AI in Games (EXAG), AIIDE 2020 Workshop}}, Virtual Event, 2020. CEUR Workshop Proceedings.

\bibitem[Koza(1994)]{Koza1994gp}
John~R Koza.
\newblock Genetic programming as a means for programming computers by natural selection.
\newblock \emph{Statistics and computing}, 4:\penalty0 87--112, 1994.

\bibitem[Bombardelli(2022)]{Bombardelli2022}
Leonardo~Boaventura Bombardelli.
\newblock \emph{Generating variations of the board game risk through evolutionary game design}.
\newblock Bachelor's thesis, Universidade Federal do Rio Grande do Sul, 2022.
\newblock URL \url{https://lume.ufrgs.br/handle/10183/252514}.

\bibitem[Boyer et~al.(2003)Boyer, Cortese, Patrignani, and Di~Battista]{boyer2003stop}
John~M Boyer, Pier~Francesco Cortese, Maurizio Patrignani, and Giuseppe Di~Battista.
\newblock {Stop minding your P’s and Q’s: Implementing a fast and simple DFS-based planarity testing and embedding algorithm}.
\newblock In \emph{International Symposium on Graph Drawing}, pages 25--36, Berlin, Heidelberg, 2003. Springer Berlin Heidelberg.

\bibitem[Browne(2008)]{browne2008automatic}
Cameron Browne.
\newblock \emph{Automatic Generation and Evaluation of Recombination Games}.
\newblock PhD thesis, Queensland University of Technology, 2008.
\newblock URL \url{https://eprints.qut.edu.au/17025/1/Cameron_Browne_Thesis.pdf}.

\bibitem[Cordella et~al.(2001)Cordella, Foggia, Sansone, Vento, et~al.]{Cordella2001isomorphism}
Luigi~Pietro Cordella, Pasquale Foggia, Carlo Sansone, Mario Vento, et~al.
\newblock An improved algorithm for matching large graphs.
\newblock In \emph{3rd IAPR-TC15 Workshop on Graph-based Representations in Pattern Recognition}, pages 149--159, Ischia, Italy, 2001. Citeseer.

\bibitem[Islam et~al.(2007)Islam, Alam, and Murase]{islam2007}
Md.~Monirul Islam, Mohammad~Shafiul Alam, and Kazuyuki Murase.
\newblock A new recurring multistage evolutionary algorithm for solving problems efficiently.
\newblock In Hujun Yin, Peter Tino, Emilio Corchado, Will Byrne, and Xin Yao, editors, \emph{Intelligent Data Engineering and Automated Learning - IDEAL 2007}, pages 97--106, Berlin, Heidelberg, 2007. Springer Berlin Heidelberg.
\newblock ISBN 978-3-540-77226-2.

\end{thebibliography}

\appendix

\section{Evaluation Metrics' Equations}
\label{equations}

The evaluation of the board for each player $p$ ($H_p$) is a heuristic that combines the number of territories and troops (Eq. \ref{heuristica}). To achieve this, weights \textit{w\textsubscript{terr}} and \textit{w\textsubscript{unit}} define the importance of territories and units, respectively, on the game. Additionally, \textit{T\textsubscript{p}} is player p's  territory count, \textit{T\textsubscript{total}} is the total territory count, \textit{U\textsubscript{p}} is player p's troop count and \textit{U\textsubscript{total}} the total troop count.

\begin{equation}
\label{heuristica}
    H_p = \frac{T_p}{T_{total}} * w_{terr} + \frac{U_p}{U_{total}} * w_{unit}
\end{equation}

Completion (\textit{C\textsubscript{comp}}) is calculated based on the number of games won by players 1 or 2, divided by the number of played games $J$ (Eq. \ref{completeness}). The best \textit{C\textsubscript{comp}} is 1 (there is always a winner). %Similar to all subsequent metrics, the overall completeness is an average of all games $J$ played in the testing phase.

\begin{equation}
\label{completeness}
    C_{comp} = \frac{wins_1 + wins_2}{J}
\end{equation}

Duration (\textit{C\textsubscript{dur}} on Eq. \ref{duration}), considers the preferred duration (\textit{D\textsubscript{pref}}) and the duration of each game $j$ played (\textit{D\textsubscript{j}}) out of $J$ games. The preferred duration was set as 24, representing 12 turns per player. %Assuming a human player takes 3 minutes per turn, this results in a game with 72 minutes. 
\textit{C\textsubscript{dur}} can range from 0 (best) to 1 (worst).

\begin{equation}
\label{duration}
    C_{dur} = \frac{1}{J} \sum_{j=1}^J{\frac{|D_{pref} - D_j|}{D_{pref}}}
\end{equation}

Advantage ($C_{adv}$ on Eq. \ref{advantage}) is determined by the win rate of player 1. $C_{adv} = 0.5$ means a balanced game, less or greater than 0.5 implies disadvantage or advantage for player 1, respectively.

\begin{equation}
\label{advantage}
    C_{adv} = \frac{|wins_1 - (wins_1 + wins_2)/2|}{(wins_1 + wins_2) / 2}
\end{equation}

The number of possible moves is measured in all three phases of the game: when the player must add new units (\textit{M\textsubscript{add}}), attack (\textit{M\textsubscript{attack}}), and move their troops (\textit{M\textsubscript{fort}}). 
\textit{M\textsubscript{add}} (Eq. \ref{add})  is the product of player p's available units \textit{u\textsubscript{p}} by its territory count \textit{T\textsubscript{p}}.

\begin{equation}
\label{add}
    M_{add} = u_p \times T_p
\end{equation}

\textit{M\textsubscript{attack}} is given by Eq. \ref{attack}. For all player p's territories \textit{T\textsubscript{p}}, the number of neighboring enemy territories \textit{T\textsubscript{fb}(terr)} is multiplied by \textit{u\textsubscript{terr}-1} troops, with 3 being the limit.

\begin{equation}
\label{attack}
    M_{attack} = \sum_{terr=1}^{T_p}{T_{fb}(terr) \times \min(3, u_{terr} - 1)}
\end{equation}

To calculate \textit{M\textsubscript{fort}} (Eq. \ref{fort}), for each territory \textit{terr} of the player, the units \textit{u\textsubscript{terr}} in \textit{terr} are multiplied by the number of neighboring territories \textit{T\textsubscript{nb}(terr)} controlled by the player. 

\begin{equation}
\label{fort}
    M_{fort} = \sum_{terr=1}^{T_p}{T_{nb}(terr) \times (u_{terr} - 1)}
\end{equation}

Then, the branching factor (\textit{C\textsubscript{bf}}) is calculated with Eq. \ref{fr}, where \textit{M(t\textsubscript{n})} = \textit{M\textsubscript{add} + M\textsubscript{attack} + M\textsubscript{fort}}, and \textit{T\textsubscript{j}} is the total number of turns played on game \textit{j}. The \textit{log\textsubscript{10}} component is divided by 2 to set a limit of nearly 100 moves. The best \textit{C\textsubscript{bf}} is 0.5, meaning 50 moves per turn on average. %, representing an average branching factor. The goal of measuring the branching factor is to at least minimize the value.

\begin{equation}
\resizebox{.9\columnwidth}{!}{$
    \label{fr}
    C_{bf} = \frac{1}{J}  \sum_{j=1}^J \min\left(1, \log_{10} \left( \frac{\sum_{n=0}^{T_j-1} M(t_n)}{T_j} + 1 \right) \Bigg/2 \right)
    $}
\end{equation}

Drama (\textit{C\textsubscript{dra}} on Eq. \ref{drama}) considers number of times the winner suffered a setback. The heuristic evaluation of the winner \textit{H\textsubscript{g}(t\textsubscript{n})} the loser \textit{H\textsubscript{p}(t\textsubscript{n})} on turn \textit{t\textsubscript{n}} are used. %Thus, we can measure the severity of each disadvantage suffered by the current winner. 
\textit{C\textsubscript{dra}} can range from 0 to 1, with the optimal value being 0.5.

\begin{equation}
\label{drama}
\resizebox{.9\columnwidth}{!}{$
    C_{dra} = \frac{1}{J}  \sum_{j=1}^J{ \left( \frac{\sum \left( H_g(t_n) < H_p(t_n) \left\{ \sqrt{H_p(t_n) - H_g(t_n)} \right\} \right)}{\text{count}(H_g(t_n) < H_p(t_n))} \right) }$
    }
\end{equation}

The killer move rate (\textit{C\textsubscript{km}}) is given by Eq \ref{me}. 
Only the largest difference between the evaluations of players ($H_a$ and $H_o$) in subsequent turns ($t$ and $t-1$) is considered for each game j out of $J$ games. %Thus, only the most extreme move is used in the calculation. 
%The heuristic evaluation of the current player \textit{H\textsubscript{a}(t\textsubscript{n})} and the opponent \textit{H\textsubscript{o}(t\textsubscript{n})} in turn \textit{t\textsubscript{n}} and the previous turn, respectively \textit{H\textsubscript{a}(t\textsubscript{n-1})} and \textit{H\textsubscript{o}(t\textsubscript{n-1})}, are used. 
\textit{C\textsubscript{km}} can range from 0 to 1, with the optimal of 0.5 (moderate amount of extreme moves).

\begin{equation}
\label{me}
\resizebox{.9\columnwidth}{!}{$
    C_{km} = \frac{1}{J} \sum_{j=1}^J \max_{1 \leq n \leq T_{j-1}} [H_a(t_n) - H_o(t_n)] - [H_a(t_{n-1}) - H_o(t_{n-1})]
    $}
\end{equation}

Lead change (\textit{C\textsubscript{lc}} on Eq. \ref{ml}) counts the times of times the leader of the current turn \textit{leader(t\textsubscript{n})} is different from the previous turn \textit{leader(t\textsubscript{n-1})}. 
Only non-final turns \textit{T\textsubscript{j}-1} of all $J$ games are considered. 
\textit{C\textsubscript{lc}} can range from 0 to 1, with the optimal of 0.5 (moderate lead change).

\begin{equation}
\label{ml}
\resizebox{.9\columnwidth}{!}{$
    C_{lc} = \frac{1}{J} \sum_{j=1}^J { \left( \frac{\text{count}(leader(t_n) \neq leader(t_{n-1}))}{T_j - 1} \right) }
    $}
\end{equation}

%%%%%%%%%%%%  Supplementary Figures  %%%%%%%%%%%%
%\clearpage

%%%%%%%%%%%%%%%%   End   %%%%%%%%%%%%%%%%
%\end{multicols}  % Method B for two-column formatting (doesn't play well with line numbers), comment out if using method A
\end{document}